\documentclass{article}

      \usepackage[preprint]{neurips_2020}

\usepackage[utf8]{inputenc} % allow utf-8 input
\usepackage[T1]{fontenc}    % use 8-bit T1 fonts
\usepackage{url}            % simple URL typesetting
\usepackage{booktabs}       % professional-quality tables
\usepackage{amsfonts}       % blackboard math symbols
\usepackage{nicefrac}       % compact symbols for 1/2, etc.
\usepackage{microtype}      % microtypography
\usepackage{amsmath,amssymb} % define this before the line numbering.
\usepackage{hyperref}       % hyperlinks
\usepackage{cleveref}
\usepackage{color}
\usepackage{multirow}
\usepackage{graphicx}
\usepackage{comment}
\usepackage{changepage}

\title{Triple Memory Networks: a Brain-Inspired Method for Continual Learning}

\author{
Liyuan Wang\textsuperscript{\rm 1,2}, Bo Lei\textsuperscript{\rm 1}, Qian Li\textsuperscript{\rm 1}, Hang Su\textsuperscript{\rm 2}, Jun Zhu\textsuperscript{\rm 2,*}, Yi Zhong\textsuperscript{\rm 1,*}\\ 
\textsuperscript{\rm 1} Tsinghua-Peking Center for Life Sciences, IDG/McGovern Institute for Brain Research, MOE\\ Key Laboratory of Protein Sciences, School of Life Sciences, Tsinghua University, Beijing, China\\
\textsuperscript{\rm 2} Dept. of Comp. Sci. \& Tech., Institute for AI, BNRist Center,\\ THBI Lab, Tsinghua University, Beijing, China\\
\textsuperscript{\rm *} Corresponding authors: dcszj@mail.tsinghua.edu.cn, zhongyithu@tsinghua.edu.cn
}

\begin{document}

\maketitle

\begin{abstract}
Continual acquisition of novel experience without interfering previously learned knowledge, i.e. continual learning, is critical for artificial neural networks, but limited by catastrophic forgetting. A neural network adjusts its parameters when learning a new task, but then fails to conduct the old tasks well. By contrast, the brain has a powerful ability to continually learn new experience without catastrophic interference. The underlying neural mechanisms possibly attribute to the interplay of hippocampus-dependent memory system and neocortex-dependent memory system, mediated by prefrontal cortex. Specifically, the two memory systems develop specialized mechanisms to consolidate information as more specific forms and more generalized forms, respectively, and complement the two forms of information in the interplay. Inspired by such brain strategy, we propose a novel approach named triple memory networks (TMNs) for continual learning. TMNs model the interplay of hippocampus, prefrontal cortex and sensory cortex (a neocortex region) as a triple-network architecture of generative adversarial networks (GAN). The input information is encoded as specific representation of the data distributions in a generator, or generalized knowledge of solving tasks in a discriminator and a classifier, with implementing appropriate brain-inspired algorithms to alleviate catastrophic forgetting in each module. Particularly, the generator replays generated data of the learned tasks to the discriminator and the classifier, both of which are implemented with a weight consolidation regularizer to complement the lost information in generation process. TMNs achieve new state-of-the-art performance on a variety of class-incremental learning benchmarks on MNIST, SVHN, CIFAR-10 and ImageNet-50, comparing with strong baseline methods. 
\end{abstract}

\section{Introduction}
The ability to continually learn new information without interfering previously learned knowledge, i.e. continual learning, is one of the basic challenges for deep neural networks (DNN), because the continual acquisition of information from dynamic data distributions generally results in catastrophic forgetting \cite{mccloskey1989catastrophic}. When accommodating for new experience, a normally trained DNN tends to adjust the learned parameters and thus forgets the old knowledge.

%\junz{the credit shouldn't only be given to the review paper. We have to refer to the original papers. Some representative CL methods should be discussed in more details, which can draw CS reviewers closer. We analyze the shortcomings of existing computational models. Then, the brain science part can be put under this context.}\lywang{Do you mean to take at least one example per category and discuss the shortcommings? I can provide theoretical analysis but cannot find direct citation. The only one to show the shortcommings is the following empirical paper  }

Numerous efforts has been devoted to mitigating catastrophic forgetting, e.g. regularization methods and memory replay \cite{parisi2019continual}. Regularization methods protect important parameters for solving the learned tasks, e.g. EWC \cite{kirkpatrick2017overcoming} and SI \cite{zenke2017continual}, but hard to allocate additional parameters for new outputs without access to old data distributions. Memory replay methods replay a small amount of training data or use deep generative models to replay generated data \cite{shin2017continual}, which usually cannot precisely maintain the distributions of the learned training data. As shown in \cite{kemker2018measuring}, the strategy that achieves the optimal performance of continual learning often heavily depends on the learning paradigms and the datasets being used, but none of the existing methods solves catastrophic forgetting. The empirical results suggest a general strategy to further alleviate catastrophic forgetting: both the two forms of information, i.e. the learned knowledge for solving old tasks and the learned distribution of old training data, should be maintained to complement the lost information with each other during continual learning.

%\junz{this sentence has no context: what two forms of information? what's consolidation or complementation? why are they of interest?}\lywang{consolidation is analoglous to protection and maintain, but a more formal term in neuroscience, is it necessary to explain it and then use it?}\junz{yes. the main criticism is on the connection between neuroscience and CS. we should make the bridge as clear as possible.}

%Regularization methods protect important synapses which encode knowledge to conduct the previously learned tasks, such as elastic weight consolidation (EWC) \cite{kirkpatrick2017overcoming} and synaptic intelligence (SI) \cite{zenke2017continual}. Memory replay methods replay specific information of the learned data distribution during training a new task, through replaying a small amount of training data or using deep generative models to replay generated data \cite{shin2017continual}. To accommodate for the specific information or generalized knowledge of the new task, progressive networks \cite{rusu2016progressive} allocates sub-networks and the hard attention (HAT) method \cite{serra2018overcoming} allocates a task-specific parameter subspace. 

\begin{figure}[t]
	\centering
	\includegraphics[width=0.9\linewidth]{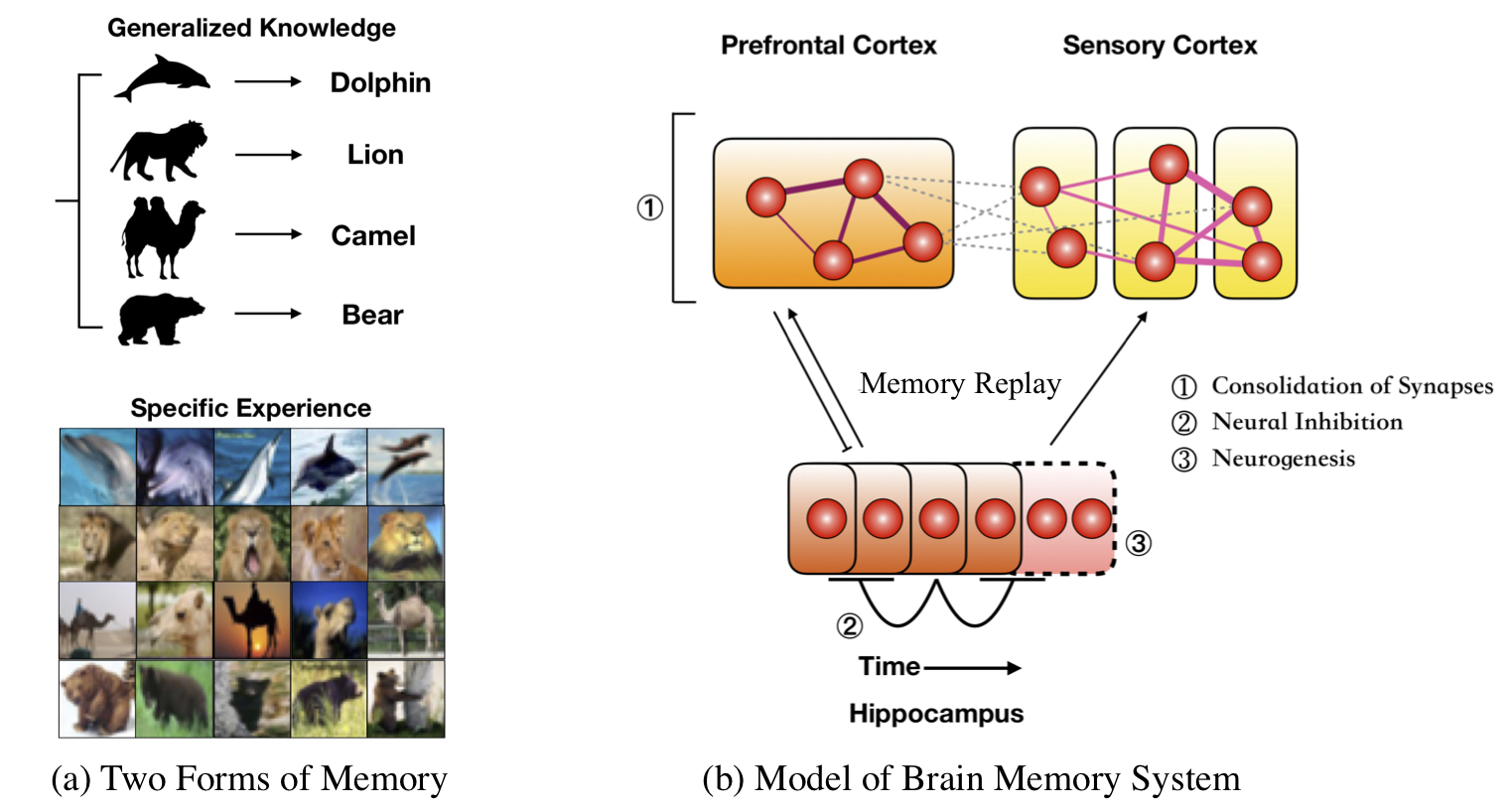}
	\caption{Diagrams of the brain memory system. Two forms of memory are encoded in three brain regions and protected by two consolidation mechanisms: 1. Consolidation of Synapses; 2. Neurogenesis \& Neural Inhibition. (b) is modified from \cite{frankland2005organization}.}
	\label{fig1}
    \vspace{-.7cm}
\end{figure}

Compared with DNNs, the brain is able to continually learn new experience without catastrophic forgetting \cite{deng2010new,wiskott2006functional,mcclelland1995there}. This ability is possibly achieved by the organization principles of the brain memory system, i.e. cooperation of the three memory networks as a currently well-accepted model \cite{frankland2005organization}. In each module of the triple-network architecture, information is encoded as more specific forms, e.g. an object you have ever seen, or more generalized forms, e.g. the features of the object that associate to a concept.
The encoded information is stabilized individually by different neural mechanisms in these modules (Fig. 1), the process of which is called \(consolidation\) of memory. Specifically, hippocampus develops the mechanisms of neurogenesis and neural inhibition to encode more specific information. Neurogenesis generates additional neurons to create space for incoming experience \cite{deng2010new}. The inhibitory neurons are developed together to inhibit irrelevant parts of the network and prevent interference \cite{gonccalves2016adult}. While, neocortex encodes more generalized information, which is consolidated by strengthening synaptic connectivity \cite{frankland2005organization}. In neocortex, prefrontal cortex (PFC) mediates the interplay of the two memory systems. PFC develops a discrimination mechanism to regulate the encoding of specific information in hippocampus and integrates (sensory) cortical modules (SC) in neocortex to encode generalized information \cite{frankland2005organization}. The interplay of hippocampus, PFC and SC complements the individually consolidated information to avoid catastrophic forgetting. This strategy inspires us to interdependently protect both the structured knowledge to solve tasks and the learned distribution of training data in a continual learning system.

%\junz{too complicated sentence! I don't understand, extremely hard to parse.}\junz{also, avoid using too strong words like "solve". We don't solve it; we just improve over previous models.}

In this work, we aim to provide a new approach for continual learning by systematically drawing inspirations from the triple-memory architecture of the brain functions. Specifically, the framework applies the idea of generative adversarial networks (GAN) \cite{goodfellow2014generative} to model the functions of hippocampus, PFC and SC as cooperation of a generator, a discriminator and a classifier, respectively. A conditional generator learns the distribution of training data, similar to the specific experience encoded in hippocampus. The generator models the role of neurogenesis and neural inhibition as an extendable architecture to learn the incoming domains and a domain-specific attention mask to prevent interference of the encoded experience. The generated data is replayed to a discriminator to model PFC, which detects familiarity of the input experience, and then to a classifier to model SC. During memory replay, the classifier learns new tasks under supervision of the discriminator, as PFC integrates SC to encode generalized information. To regularize the difference between the generated data and the old training data, both the classifier and the discriminator are implemented with weight consolidation algorithms, inspired by the strengthened synaptic connectivity in neocortex. The model is named as ``triple memory networks (TMNs)'' (Fig. 2), since the three networks continually learn the knowledge for generation, identification and classification and maintain the learned knowledge by corresponding brain-inspired algorithms.

Our contributions include: (1) We propose that the brain strategy, processing both specific and generalized information by interplay of three memory networks, can be applied in continual learning; (2) We present the triple memory networks (TMNs), an ``artificial memory system'' for generative replay and class-incremental learning, which leverages the organization principles of brain memory system to mitigate catastrophic forgetting; (3) Our method achieves  state-of-the-art (SOTA) performance on a variety of image classification benchmarks, including MNIST, SVHN, CIFAR-10 and ImageNet-50; and (4) To the best of our knowledge, TMNs is the first attempt to model the triple-network architecture of brain memory system for continual learning, which bridges the two fields of artificial neural networks and biological neural networks.

\section{Related Work}
Regularization methods use an additional regularization term in the loss function to penalize changes of important parameters for the old tasks and thus stabilize these weights (synapses).  \cite{kirkpatrick2017overcoming} proposed the elastic weight consolidation (EWC) method, which adds a quadratic penalty on the difference between the parameters for the old and the new tasks. In EWC, the strengths of penalty (the ``importance'') on each parameter are calculated by the diagonal of the Fisher Information Matrix (FIM). \cite{zenke2017continual} proposed the synaptic intelligence (SI) approach to calculate synaptic relevance in an online fashion. Memory aware synapses (MAS) \cite{aljundi2018memory} applied gradient of the learned function with respect to the parameter as the importance of the synapse. Interestingly, \cite{aljundi2018memory} showed that the importance of a synapse is equal to the co-activation frequency of the two neurons connected by the synapse in MAS. This principle is analogous to the Hebbian learning theory of synaptic plasticity: ``Cells that fire together, wire together'' \cite{hebb1962organization}.  \cite{serra2018overcoming} proposed a hard attention to the task (HAT) algorithm to allocate dedicated parameter subspace for individual tasks by a hard attention mask, similar to the function of biological neural inhibition. 

Memory replay strategies mitigate catastrophic forgetting by replaying a small amount of training data or replaying generated data (generative memory replay) through deep generative model. \cite{rebuffi2017icarl} proposed the iCarl approach to store example data points of old tasks to train the feature extractor together with new data. EEIL used a distillation measure to retain the exemplar set of old classes \cite{castro2018end}. However, in the strict continual learning setups, to store real data is not allowed. Generative memory replay avoids the limitation of replaying raw data by reconstruction of training examples from a generative model. FearNet \cite{kemker2017fearnet} applied a generative auto-encoder to train a classifier together with training data. Deep Generative Replay \cite{shin2017continual}, Memory Replay GAN (MeRGAN) \cite{wu2018memory} and Dynamic Generative Memory (DGMw) \cite{ostapenko2019learning} used GAN as the generative model. 

To adapt the setup of incremental tasks, some continual learning frameworks make network extendable. \cite{rusu2016progressive} proposed a progressive network to freeze the network trained on previous tasks and allocate new sub-networks to train incoming tasks. Several generative replay methods make the generative network extendable to learn new domains, such as FearNet, MeRGAN and DGMw. 

Performance of continual learning is mainly evaluated on task-incremental or class-incremental scenarios \cite{vandeven2019three,ostapenko2019learning}. In task-incremental scenarios, the task label is given at test time, i.e. multiple-head evaluation. The setups of class-incremental learning, which predominately applies single-head evaluation \cite{chaudhry2018riemannian}, is more general and realistic, that the task label is not provided during testing. However, single-head evaluation usually requires the model to allocate dedicated parameters for new outputs of the incremental classes to evaluate them together. To allocate additional parameters under iid data distributions is challenging for weight regularization methods without access to training data. Sub-network methods with attention gate decorrelate the learned tasks and thus fails to evaluate them together. Memory replay show advantages in such incremental setting \cite{kemker2018measuring} but limited by unavailability to the real data. Generative memory replay is a promising strategy which applies generated data to replace the unavailable old training data, but only perform well on relatively simple datasets, e.g. MNIST and SVHN \cite{parisi2019continual}.

Our method aims to improve continual learning on class-incremental setups and single-head evaluation, which can be easily extended to other scenarios. Thus, we analyze two main issues of existing generative replay methods and solve these issues through brain-inspired strategies, as shown below.

\section{Preliminary Knowledge}

We start by introducing the notations and the setup of continual learning and discuss on the limitations of existing methods. 

\subsection{Setups and Notations}

We consider the continual learning with $T$ tasks. The entire dataset \( S = \bigcup_{t\in[T]} S_{t}\) is a union of task-specific datasets \(S_{t} = \{(x_{i}^{t}, y_{i}^{t})\}_{i=1}^{{N}_{t}} \), where $[T]=\{1, \cdots, T\}$. \(x_{i}^{t} \in X \) is an input data with the true label \(y_{i}^{t} \in Y\). In the incremental setups, the dataset \(S_{t}\) is only available during training task \(t\). \(S_{t}\) contains the data from multiple classes or only a single class. The testing part follows the single-head evaluation, i.e. task labels are not provided and all the learned classes \(Y\) are evaluated at test time \cite{chaudhry2018riemannian}.

In this setting, suppose a DNN first learns the parameters from the dataset \(S_{t}\) of the current task \(t\) and can perform well on it. For the new task $t+1$, if the DNN is directly trained on \(S_{t+1} \) without the availability of \(S_{t}\), the parameters will adapt to achieve good performance on task \(t+1\), while tend to forget the learned knowledge of task \(t\). This is known as catastrophic forgetting~\cite{mccloskey1989catastrophic}.

\subsection{Two Issues of Existing Generative Replay Methods}
To avoid forgetting information of old tasks in continual learning, a promising strategy is {\it generative replay}, which first learns a generative model to describe the old training data and then replays the model to generate data \(\hat{S}_{<t}\) of all the previous tasks. When training task $t$, we use the extendable dataset \(S' = S_{t} \bigcup\hat{S}_{<t}\), which contains both the training data \(S_{t}\) of the current task and \(\hat{S}_{<t}\), thereby preventing the forgetting of previous knowledge. In the sequel, we use \(c\) to denote the label of a generated data, distinguishing from the training data label \(y\). Because \(S'\) consists of both training data \(S_{t}\) and generated data \(\hat{S}_{<t}\), we use \(y_{c}\) to denote the label of \(x\) in \(S'\). 

However, current state-of-the-art (SOTA) generative replay methods, such as MeRGAN \cite{wu2018memory} and DGMw \cite{ostapenko2019learning}, only perform well on relatively simple datasets (e.g., MNIST and SVHN) while much poorer in complex datasets (e.g., ImageNet). Here we show two issues of current generative replay methods which decrease performance of continual learning:

\textbf{Issue (a): Difference between the generated data and training data.} Generative replay does not directly solve catastrophic forgetting, but transfers the stress from the task-solver network to the generative network. Catastrophic forgetting in generative replay methods is mainly caused by the deviance between generated data and training data, because if the generative module precisely learns the distribution of old training data, the performance of generative replay should be the same as joint training of all training data. To show the difference of generated data from training data on the task-solver network during generative replay, we train a classifier first with training data to simulate the training stage, and then with generated data of the same task to simulate the replay stage in a 10-class ImageNet task. Then we measure the empirical Fisher information matrix (FIM), i.e., squared gradients of the parameters, on training data and generated data, and quantify the cosine similarity in Table 3 (\(\lambda_{C}=0\) group). Cosine similarity measures the cosine of angle between the two matrices through the normalized inner product. The divergent directions of the two empirical FIMs indicate that the parameters of the classifier are optimized to different directions on the two types of data (i.e., generated data and training data).

%\junz{can you put this slightly more formal? Not sure about how some key values are calculated, e.g., cosine similarity between two FIM?}
%This issue can be solved by weight regularizatio on classifier in our design.\junz{defer this to where our method is presented, and keep the analysis of the "issues" clear.}

\textbf{Issue (b): Interference of discrimination and classification in a joint discriminator network.} Many SOTA generative replay methods are implemented in a two-module architecture of GANs, e.g., AC-GAN~\cite{odena2017conditional}, where the discriminator network is responsible for both discriminating fake/real examples and predicting class labels. However, in continual learning, the discriminator is optimized for discrimination of the current task \(t\) on dataset \(S_{t}\) and classification of all the learned tasks \(\le t\) on dataset \(S' = S_{t} \bigcup\hat{S}_{<t}\). Thus, discrimination and classification in a joint discriminator network might not achieve optimum simultaneously. To verify this interference, we measure empirical FIM of the auxiliary classifier and the discriminator after training an AC-GAN model continually with 10 classes of MNIST or SVHN. The two FIMs show poor overlap in the shared part of the discrimination network (Table 4). Particularly, the weight of deeper convolution layers shows much poorer similarity. The averaged cosine similarity of the conv1.weight, conv2.weight and conv3.weight are 0.66, 0.40, 0.16 in MNIST and 0.70, 0.20, 0.08 in SVHN. Thus, we hypothesize that discrimination and classification interferes with each other in the continual learning setups.
%\junz{which two functions? You may use necessary notations to again put the arguments more formal and precise.} 

\section{Our Proposal}
%\subsection{Preliminary Analysis}
We now present our approach to addressing the above issues of existing memory replay methods. We build on the success of brain memory system in continual learning and develop three ``artificial memory networks'' with implementing consolidation approaches in appropriate modules. 

\begin{figure}[t]
	\centering
	\includegraphics[width=0.8\linewidth]{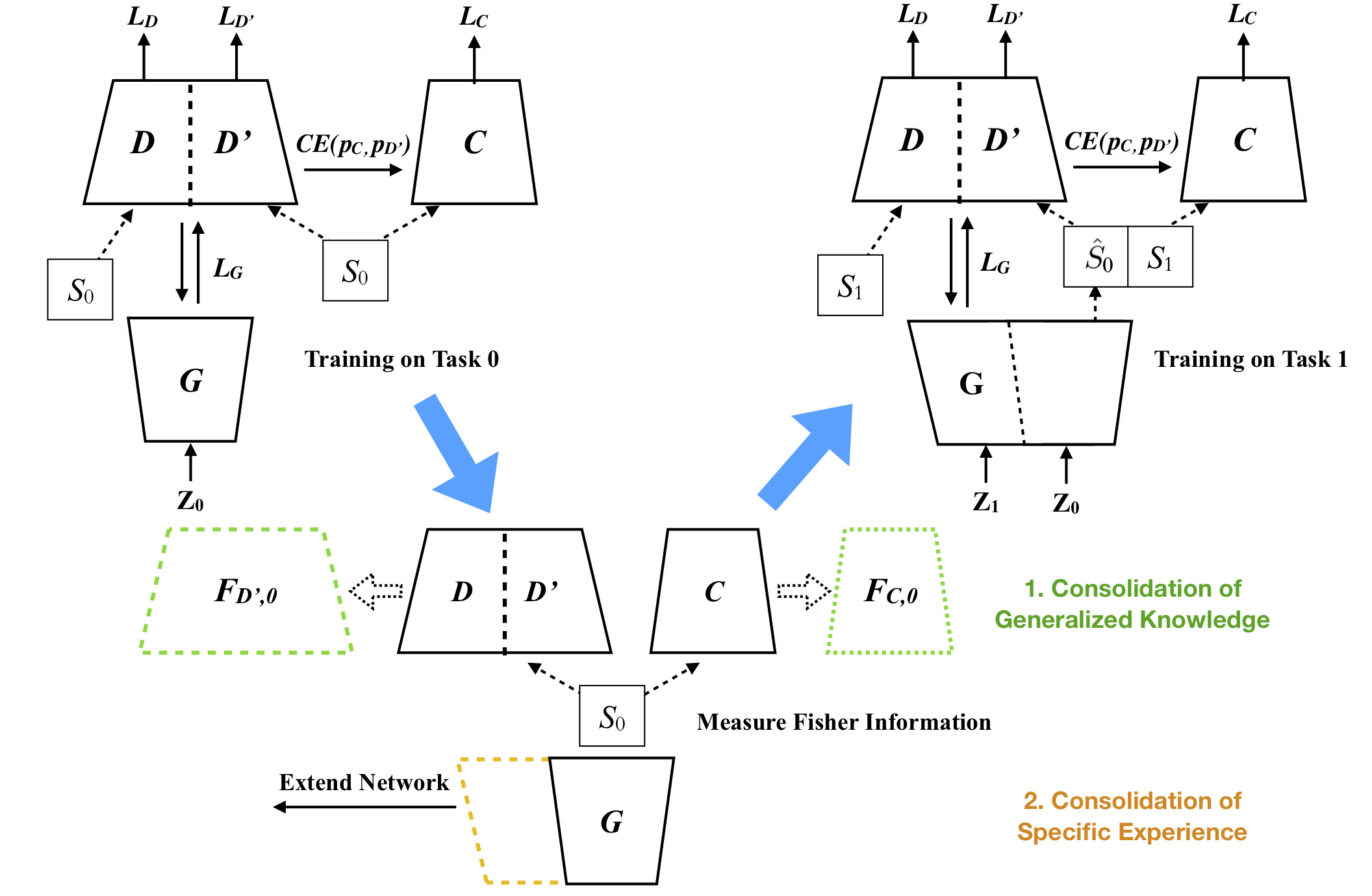}
	\caption{Architecture and Training of TMNs. \textit{G}, \textit{D}, \textit{D'}, \textit{C} represent generator, discriminator, auxiliary classifier and classifier, respectively. \(S_{t}\) and \(\hat{S}_{t}\) is training and generated dataset of task \(t\).}
	\label{fig2}
    \vspace{-0.5cm}
\end{figure}
\subsection{Triple Memory Networks}

%\junz{We can start with an over-view of the TMN network for CL and present its main components, and use subsections/paragraphs to introduce each component in detail as well as their interplay... }\lywang{I organize following paragraphs as: paragraph1 to introduce the base model of hippocampus-PFC for generative replay, which is technically very similar to the current SOTA method (DGMw); then paragraph2 introduce "the third network", i.e. classifier, to solve issue (b); then paragraph3 intruduce "weight consolidation", to solve issue (a). I think this organization will make our technical novelty more clear to the people who is familiar with current SOTA method.}
Fig. 2 illustrates the overall architecture of our Triple Memory Networks (TMNs), which consists of a generator ($G$), a discriminator ($D$) and a classifier ($C$). Before diving into the implementation details, we provide rationale on the design of each module. %\junz{give the structure first, then refer to it when explaining.}

Because of the similar function of the generative module and hippocampus to reconstruct specific information of the learned experience, we start with modeling the mechanisms of hippocampus in $G$ to mitigate catastrophic forgetting in generation process. Hippocampus accommodates for new information without interfering previous experience by neurogenesis and neural inhibition \cite{deng2010new}, and it continually generates new-born neurons to encode new experience and develop inhibitory neurons to inhibit irrelevant parts of the network. These neurons quickly become mature and decrease plasticity \cite{deng2010new}. Thus, even when the incoming pattern is similar to a previous one, they will not interfere with each other \cite{rolls2013mechanisms}. The hippocampal neurogenesis can be modeled as an extendable generative network and the neural inhibition is close to a binary attention mechanism \cite{serra2018overcoming}. Since PFC inhibits hippocampus to prevent encoding of redundant information when the incoming experience matches a previously learned cortical memory \cite{frankland2005organization}, we model the function of PFC as a discriminator and the hippocampus-PFC interaction as adversarial training of GAN. Such a combination has been proved effective to generate images in incremental simple domains \cite{ostapenko2019learning}, although performs much poorer in complex datasets (e.g., ImageNet).

To address the above issue (b), we use a relatively independent classifier \(C\), which is optimized only for classification rather than two objectives, inspired by organization principles of neocortex. Generalized information (e.g., structured knowledge for discrimination and classification) is encoded in neocortex but supported by different regions. In particular, sensory cortex (SC) is the dedicated region to extract features of sensory input, e.g., visual information \cite{frankland2005organization} and maintain generalized sensory information, analogous to the function of a classifier. 

To address the above issue (a), we use a weight consolidation regularizer based on empirical FIM on classification process to regularize the divergence of training data and generated data during generative replay. This strategy is inspired by the strengthened synaptic connections in neocortex in biological memory replay. During biological memthe replay, specific information encoded in hippocampus is transferred into PFC and SC as generalized knowledge, corresponding to the knowledge of discrimination and classification in our framework. To consolidate the generalized knowledge, neocortical regions (e.g., PFC and SC) incrementally strengthen synaptic connections to stabilize the synapses connecting simultaneously-activated neurons. Particularly, our weight regularization process keeps biologically plausible because empirical FIM is also a measurement of synaptic connectivity of a network \cite{achille2017critical}.

%During biological memory replay, the specific experience encoded in hippocampus is transferred into PFC and SC as generalized knowledge, corresponding to the knowledge of discrimination and classification in our framework. To consolidate the generalized knowledge, neocortical regions (e.g., PFC and SC) incrementally strengthen synaptic connections to stabilize the synapses connecting simultaneously-activated neurons.  Particularly, the weight regularization process is biologically plausible because empirical FIM is also a measurement of synaptic connectivity of a network \cite{achille2017critical}.

Note that the closest GAN architecture to TMNs is Triple-GAN \cite{chongxuan2017triple}, which also applies an additional classifier. But Triple-GAN is proposed to improve classification and class-conditional image generation in semi-supervised learning. So the classifier is trained to label the unlabeled data. In contrast, TMNs is designed to alleviate catastrophic forgetting in continual learning, thus the three networks are implemented with consolidation algorithms inspired by the brain memory system, as detailed below.

\subsection{Implementation Details}
We now present the implementation details of TMNs with the interaction of \(C\) with \(D\) and \(G\). In biological memory replay, SC and other cortical modules learn generalized information under the integration of PFC. During training task \(t\), \(G\) randomly generates data \(x\) with label \(c\) using a random noise vector \(z\) of the current task \(t\) and all the learned tasks \(0,...,t-1\): The sampling distribution of $c$ is uniform \({p}_{c} = \textit{U}\{ 1, t\}\) and \(z\) is Gaussian \(p_{z} = N(0, 1)\). After training each task \(t\), \(G\) generates the dataset \(\hat{S}_{t} = \{(x_{i}^{t}, c_{i}^{t})\}_{i=1}^{{N}_{t}}\) to update \(S'\). To model the interaction of PFC and SC, although both \(D\) and \(C\) should receive the replay dataset \(S'\), \(D\) supervises \(C\) to learn the generalized knowledge. Since \(C\) learns \(p_{C}(y|x)\) to classify the data, \(D\) should learn not only \(p_{D}(x)\) to identify fake data but also \(p_{D}(y|x)\). 

A straightforward design is to use \(D\) to model \(p_{D}(x)\) and an auxiliary classifier \(D'\) to learn \(p_{D'}(y|x)\). So the optimization problem becomes to optimize four groups of parameters in three networks: \{\({\theta^{D},\theta^{ D'},\theta^{ C},\theta^{ G}}\)\}. To stabilize the training process, the loss function of the discriminator \(L_{D}\) follows WGAN-GP \cite{gulrajani2017improved}. The loss function of the auxiliary classifier \(L_{D'}\) consists of a cross entropy term and an elastic weight consolidation (EWC) regularizer on empirical FIM \({F}_{D'}\).
\begin{equation} {L}_{D'}(\theta^{ D'}) = {L}_{CE}(\theta^{ D'}) +  \lambda _{D'}\sum_{i} {F}_{D',i}( \theta^{ D'} _{i } -  \theta^{ D'})^{2},  \end{equation} where the cross entropy \(L_{CE}(\theta^{ D'}) \) is calculated from the classification results of the auxiliary classifier \(p_{D'}(y|x)\) and the true labels on training data of current task and generated data of previous tasks:
\begin{equation}
{L}_{CE}(\theta^{ D'}) = - \mathbb{E}_{(x,\, y_{c})\sim S'}\,[y_{c} \, \log\,D'(x)].
\end{equation} 
For notation clarity, we have used $D'(x)$ to denote $D'_{ \theta^{ D'}}(x)$ without explicitly writing out its parameters, likewise for $D$, $G$ and $C$.

Because of the deviance between the generated data and the training data of the previous tasks, only to minimize \(L_{CE}(\theta^{ D'}) \) cannot optimize \(p_{D'}(y|x)\) the same as the true distribution \(p_{l}(y|x)\). In theory, the gap can be filled by the regularization term since the \({F}_{D'} \) is directly calculated from training data. In class-incremental setups, the output layer expands once a dimension for each incremental class and changes the shape of its weight. The regularization term only calculates and protects the parameters of other layers except for the output layer, i.e. the shared network of \(D\) and \(D'\). The loss function \(L_{D} \) is:

\begin{equation}  
\begin{split}
{L}_{D}(\theta^{ D}) =  & -\mathbb{E}_{(x, \, y_{c})\sim S'}\,[D(x)] +  \mathbb{E}_{c\sim p_{c},  z \sim p_{z}}\,[D(G(z, c))]  \\
& + \lambda _{GP}\,\mathbb{E}_{(x,\, y_{c}) \sim S'}\,[ (\left\|\nabla D(x)\right\| _{2} - 1)^{2}].
\end{split}
\end{equation} 
     
The loss function of the classifier \(L_{C}\) in (4) includes a cross entropy term and an regularization term. The cross entropy term \({CE}_{\theta^{C}}(p_{C}(y|x),p_{D'}(y|x))\) minimizes the difference of \(p_{C}(y|x)\) and \(p_{D'}(y|x)\) on the replay dataset \(S'\) to transfer knowledge from \(D'\) to \( C\), since the regularization term in (2) has penalized the gap between \(p_{D'}(y|x)\) and \(p_{l}(y|x)\).  Similar to \(D'\), the \(p_{C}(y|x)\) and \(p_{D'}(y|x)\) on previous tasks are calculated from \(S'\) and the gap of imperfect generation can be filled by the weight consolidation regularizer. The Fisher Information \({F}_{C}\) of \(C\) is not directly calculated from its loss function \({L}_{C}\) but \(L_{C'}\), including an additional cross-entropy \({L}_{ CE}(\theta^{ C})\)  of the classification results \(p_{C}(y|x)\) and the ground truth labels of training data \(p_{l}(y|x)\) to minimize the gap in the two distributions.

\begin{equation} 
\begin{split}
{L}_{C}( \theta^{ C} ) =
-\mathbb{E}_{(x, \, y_{c})\sim S' \,}[C(x) \, \log\,D'(x)] +
\lambda _{C} \sum_{i} {F}_{C,i}( \theta^{ C} _{i } -  \theta^{ C})^{2}, 
\end{split}
\end{equation} 
\begin{equation} {L}_{C'}( \theta^{ C}) = {L}_{ CE}(\theta^{ C}) + {L}_{C}(\theta^{ C}), \end{equation} 
\begin{equation}
{L}_{CE}(\theta^{ C}) = - \mathbb{E}_{(x,\, y_{c})\sim S'}\,[y_{c} \, \log\,C(x)].
\end{equation} 

The conditional generator \(G\) uses an extendable network and the hard attention masks to model hippocampus. We apply a similar conditional generator architecture as DGMw \cite{ostapenko2019learning}: \(m_{t}^{l} = \sigma(s \cdot e_{t}^{l})\) is the attention weight of the layer \(l\) at task \(t\) with the initialization of \(0.5\), where \(s\) is a positive scaling factor, \(e_{t}^{l}\) is a mask embedding matrix and \(\sigma\) is the sigmoid function. To prevent interference of the learned generation tasks, the gradients \(g_l\) of each layer \(l\) is multiplied by the reverse of cumulated attention mask \(m_{\leq t}^{l}\) of the learned tasks \(\leq t\):
\begin{equation}
g_{l} = (1 - m_{\leq t}^{l})  \,g_{l} ,  \,m_{\leq t}^{l} = max[m_{t}^{l}, m_{t-1}^{l}].
\end{equation} 

The loss function of the generator also follows the requirements of WGAN-GP and \begin{math}{R}_{M} \end{math} is the sparsity regularizer of the attention masks, where \(N_{l}\) is the number of parameters of layer \(l\):

\begin{equation} 
\begin{split}
{L}_{G}(\theta^{ G}) = 
- \mathbb{E}_{c\sim p_{c},  z \sim p_{z}}\,[D(G(z, c))] + R_{M}
-\lambda _{G} \mathbb{E}_{c\sim p_{c},  z \sim p_{z}}\,[c \, \log\,D'(G(z, c))],
\end{split}
\end{equation} 

\begin{equation} 
R_{M} = \dfrac{\sum_{l}\sum_{i=1}^{N_{l}}m_{t,i}^{l}(1-m_{<t,i}^{l})}{\sum_{l}\sum_{i=1}^{N_{l}}(1-m_{<t,i}^{l})}.
\end{equation} 

Because both \(D'\) and \(C\) can make a prediction, a decision-making process is required to decide which classifier to use. Here, we adopt a simple yet effective decision-making rule. Specifically, \(D'\) and \(C\) first estimate the probability \(P_{D'}(y=k|x)\) and \(P_{C}(y=k|x)\) that an input feature vector \(x\) belongs to class \(k\) out of all \(K\) classes, where $[K]=\{0,1, \dots, K-1\}$. Then the final prediction is made by the network with the highest confidence:
\begin{equation} \hat{y} = \mathop{\arg\max}_{k \in [K] }(  \mathop{\max}[P_{D'}(y=k|x), P_{C}(y=k|x)]).\end{equation}

%\subsection{TMNs and Brain Memory System}

%To help understand the brain-inspired system, here we discuss connections of our TMNs and the brain memory system at three levels (Fig. 1, 2):
%\begin{itemize}
 %   \item \textbf{General Functions}: The specific experience encoded in hippocampus and the generalized knowledge encoded in neocortex (PFC and SC) are respectively analogous to the generated data in \(G\) and the learned knowledge for identification and classification in \(D\) and \(C\). PFC and SC are also responsible for detection of novelty and feature extraction respectively, similar to the functions of \(D\) and \(C\).
%    \item \textbf{Relations of Networks}: During memory replay, the specific experience reconstructed from hippocampus is transferred into neocortex, where PFC integrates SC to encode the generalized knowledge; the generated data in \(G\) is replayed to \(D\) and \(C\), where C learns the knowledge of classification under the supervision of \(D\). Similar to the role of \(D\) in adversarial training of GAN, when the input experience matches a previously learned memory, PFC inhibits hippocampal activity to prevent encoding of redundant information.
%    \item \textbf{Consolidation Mechanisms}: Hippocampus prevents interference of the encoded experience through neurogenesis and neural inhibition, similar to the extendable network and the attention mechanism in \(G\); Neocortex (PFC and SC) stabilizes the learned knowledge through strengthening synaptic connections, close to the weight consolidation in \(D\) and \(C\).
%\end{itemize}

\section{Experiment}
\subsection{Experiment Setup}

Our framework is evaluated following the class-incremental setups on four benchmark datasets: MNIST \cite{lecun1998mnist}, SVHN \cite{netzer2011reading}, CIFAR-10 \cite{krizhevsky2009learning} and ImageNet \cite{russakovsky2015imagenet}. The evaluation measure is the average accuracy ($ A_{t} $) on the test set of the class $0, ..., t$ trained so far.

{\bf Datasets:} MNIST includes 50,000 training samples, 10,000 validation samples and 10,000 testing samples of black and white handwritten digits of size 28 $ \times $ 28. SVHN includes 73,257 training samples and 26,032 testing samples and each is a colored digit in various environments of size 32 $ \times $ 32. CIFAR-10 contains 50,000 training samples and 10,000 testing samples of 10-class colored images of size 32 $ \times $ 32. iILSVRC-2012 dataset contains 1000 classes of images and 1300 samples per class. We randomly choose 50 classes of iILSVRC-2012 as a subset ImageNet-50 and resize all images to 32 $ \times $ 32 before experiment. The 50 classes of images in ImageNet-50 are trained with incremental bach of 10. We use top-1 and top-5 accuracy as the evaluation measure of ImageNet-50 on the val part of iILSVRC-2012. The data shown in Table 1, 2, 5 use top-1 accuracy. All the experimental results are averaged by 10 runs.

{\bf Architecture:} We apply a 3-layer DCGAN architecture \cite{radford2015unsupervised} for the MNIST, SVHN and CIFAR-10 experiments and a ResNet-18 architecture for the ImageNet-50 experiment. The discriminator and the classifier use similar architecture except for the output layer. The generator applies an extendable network with hard attention masks similar to \cite{ostapenko2019learning}.

{\bf Baselines:} We primarily compare with the continual learning methods following the strict setups, i.e. without storing training samples. In particular, because our method is basically a generative memory replay approach, we compare with other methods dependent on the similar idea. To best of our knowledge, DGMw \cite{ostapenko2019learning} achieves state-of-the-art (SOTA) performance of class-incremental learning in most benchmark datasets, followed by MeRGAN \cite{wu2018memory}, DGR \cite{shin2017continual} and EWC-M \cite{seff2017continual}. We compare our results directly with DGMw, under the same architecture and hyperparameters for the fair comparison. In ImageNet-50 experiment, we also compare with iCarl \cite{rebuffi2017icarl} and EEIL \cite{castro2018end}, the SOTA methods to incrementally learn complex domains but have to store training samples. Since the relaxed DGMw (DGMw-R) outperforms iCarl on $ A_{30} $ when accessible to training samples, we also compare our method with DGMw-R. iCarl, EEIL and DGMw-R are allowed to keep (total number of classes in the dataset) \(\times\) (20 training samples per class). All the experiments apply joint training as the upper bound performance.

To examine the idea of weight consolidation, we apply SVHN benchmark to evaluate our system implemented with EWC or SI \cite{zenke2017continual}, another method to incrementally stabilize important parameters. SI uses an additional quadratic surrogate loss to replace the EWC term in loss functions.  To calculate the synaptic relevance in SI, we also use the same loss functions of the classifier and the discriminator as EWC. 

\subsection{Comparison with SOTA Methods }
The quantification of the comparison experiment with other methods is summarized in Table 1. We apply joint training as the upper bound performance. We compare the averaged accuracy of 5-class ($ A_{5} $) and 10-class ($ A_{10} $) in MNIST, SVHN and CIFAR-10, 30-class ($ A_{30} $) and 50-class ($ A_{50} $) in ImageNet-50. Our method outperforms the SOTA methods in SVHN and achieves comparable results in MNIST. Our method also achieves SOTA performance on CIFAR-10. Particularly, our experiment results in Table 2 show that our method more significantly outperforms the SOTA method on both $ A_{5} $ and $ A_{10} $. The difference is possibly caused by the different network architectures from \cite{ostapenko2019learning}. Our approach significantly outperforms DGMw on ImageNet-50 (Fig. 3). When accessible to the real training samples, iCarl, EEIL and DGMw-R are the SOTA methods on ImageNet. TMNs outperforms the three methods or achieves comparable results on the ImageNet-50 benckmark, although stores no training samples. 

\newcommand{\tabincell}[2]{\begin{tabular}{@{}#1@{}}#2\end{tabular}} 
\begin{table*}[ht]
	\centering
	\caption{Averaged accuracy (\%) of class-incremental learning on image. The results of baselines are cited from \cite{wu2018memory,chaudhry2018riemannian,ostapenko2019learning}.}\smallskip
	\resizebox{1\textwidth}{!}{ % If your table exceeds the column or page width, use this command to reduce it slightly
	\begin{tabular}{cccccccccc}
		\hline
		\multicolumn{2}{c}{} & \multicolumn{2}{c}{MNIST} & \multicolumn{2}{c}{SVHN} & \multicolumn{2}{c}{CIFAR-10} & \multicolumn{2}{c}{ImageNet-50}\\
		&Methods & $ A_{5} $ & $ A_{10} $ & $ A_{5} $ &$ A_{10} $ & $ A_{5} $  & $ A_{10} $ & $ A_{30} $ & $ A_{50} $\\
		\hline
		&Joint Training&99.87&99.24&92.99&88.72&83.40&77.82&57.35&49.88\\
		\hline
		\multirow{3}*{\tabincell{c}{+ Training\\ Data}}
		&EEIL (\cite{castro2018end}) & - & - &-&-& - & - & 27.87 & 11.80\\
		&iCarl (\cite{rebuffi2017icarl}) & 84.61 & 55.8 &-&-& 57.30 & 43.69 & 29.38 & \textbf{28.98}\\
		&DGMw-R (\cite{ostapenko2019learning}) & - & -& - & - &- & - & 36.87 & 18.84\\
        \hline
		\multirow{5}*{\tabincell{c}{- Training\\ Data}}
		&EWC-M (\cite{seff2017continual}) & 70.62 & 77.03 & 39.84 & 33.02 &-&-&-&-\\
		&DGR (\cite{shin2017continual}) & 90.39 & 85.40 & 61.29 & 47.28 &-&-&-&-\\
		&MeRGAN (\cite{wu2018memory}) & 98.19 & \textbf{97.00} & 80.90 & 66.78 &-&-&-&-\\
		&DGMw (\cite{ostapenko2019learning}) & 98.75 & 96.46 & 83.93 & 74.38 & 72.45 & 56.21 & 32.14 & 17.82\\
		&TMNs (ours) & \textbf{98.80} & 96.72 & \textbf{87.12} & \textbf{77.08} &\textbf{ 72.72} & \textbf{61.24} &\textbf{ 38.23 }& 28.08\\
		\hline
	\end{tabular}
	}
	\label{table1}
	\vspace{-.2cm}
\end{table*}

\begin{table*}[ht]
	\centering
	\vspace{-.1cm}
	\caption{Averaged accuracy (\%) ($ \pm $SEM)
	on SVHN, CIFAR-10 and ImageNet-50, averaged by ten runs. *The performance of DGMw is our results.}\smallskip
    %\vspace{-.2cm}
	\resizebox{1\textwidth}{!}{ % If your table exceeds the column or page width, use this command to reduce it slightly
	\begin{tabular}{ccccccc}
		\hline
		\multicolumn{1}{c}{}& \multicolumn{2}{c}{SVHN} & \multicolumn{2}{c}{CIFAR-10} & \multicolumn{2}{c}{ImageNet-50}\\
		Methods & $ A_{5} $ &$ A_{10} $ & $ A_{5} $  & $ A_{10} $ & $ A_{30} $ & $ A_{50} $  \\
		\hline
		DGMw* (\cite{ostapenko2019learning}) &84.82($ \pm $0.30)&73.27($ \pm $0.35)&68.85($ \pm $0.25)&54.40($ \pm $0.65)&30.34($ \pm $0.63)&17.84($ \pm $0.44)\\
        TMNs (w/o EWC) &84.99($ \pm $0.36)&73.81($ \pm $0.36)&69.18($ \pm $0.49)&55.70($ \pm $0.48)&32.05($ \pm $ 0.63)&18.18($ \pm $0.42)\\
		TMNs (D'+EWC) &86.56($ \pm $0.35)&75.28($ \pm $0.42)&71.16($ \pm $0.36)&59.80($ \pm $0.21)&36.05($ \pm $ 0.89)&25.36($ \pm $0.59)\\
		TMNs (C+EWC) & 86.34($ \pm $0.31)&75.45($ \pm $0.26)&70.33($ \pm $0.22)&57.32($ \pm $0.44)&34.21($ \pm $0.60)&19.55($ \pm $0.49)\\
		TMNs (C, D'+EWC) &\textbf{87.12($ \pm $0.22)}&\textbf{77.08($ \pm $0.26)}&\textbf{72.72($ \pm $0.36)}&\textbf{61.24($ \pm $0.14)}&\textbf{38.23($ \pm $0.75)}&\textbf{28.08($ \pm $0.33)}\\
		TMNs (C+SI)&86.29($ \pm $0.35)&75.40($ \pm $0.34)&-&-&-&-\\
		TMNs (C, D'+SI)&86.41($ \pm $0.27)&75.58($ \pm $0.20)&-&-&-&-\\
		\hline
	\end{tabular}
	}
	\label{table2}
	\vspace{-.2cm}
\end{table*}

\subsection{Effectiveness of Weight Consolidation}
Next, three evidences support the effectiveness of weight consolidation algorithms in the framework. The first evidence is the parallel experiment of SI and EWC, two comparable methods to approximate synaptic relevance \cite{parisi2019continual}. We implement SI into our system in the same way as EWC and compare the performance of the two models in SVHN dataset (Table 2). Implementation of EWC or SI into only the classifier or both the classifier and the discriminator results in a similar improvement of the averaged accuracy. 

The second evidence is to compare the model implemented with EWC on both \(D'\) and \(C\), or only one network, or without EWC in Table 2. In all experiments, EWC implemented in both \(D'\) and \(C\) outperforms EWC in a single network and TMNs w/o EWC. TMNs implemented with EWC in a single network also outperforms TMNs w/o EWC. Particularly, TMNs with EWC on the two classifiers significantly outperform TMNs w/o EWC on ImageNet-50 (Fig. 3).

\begin{figure}[h]
    \vspace{-.2cm}
	\centering
	\includegraphics[width=0.6\columnwidth]{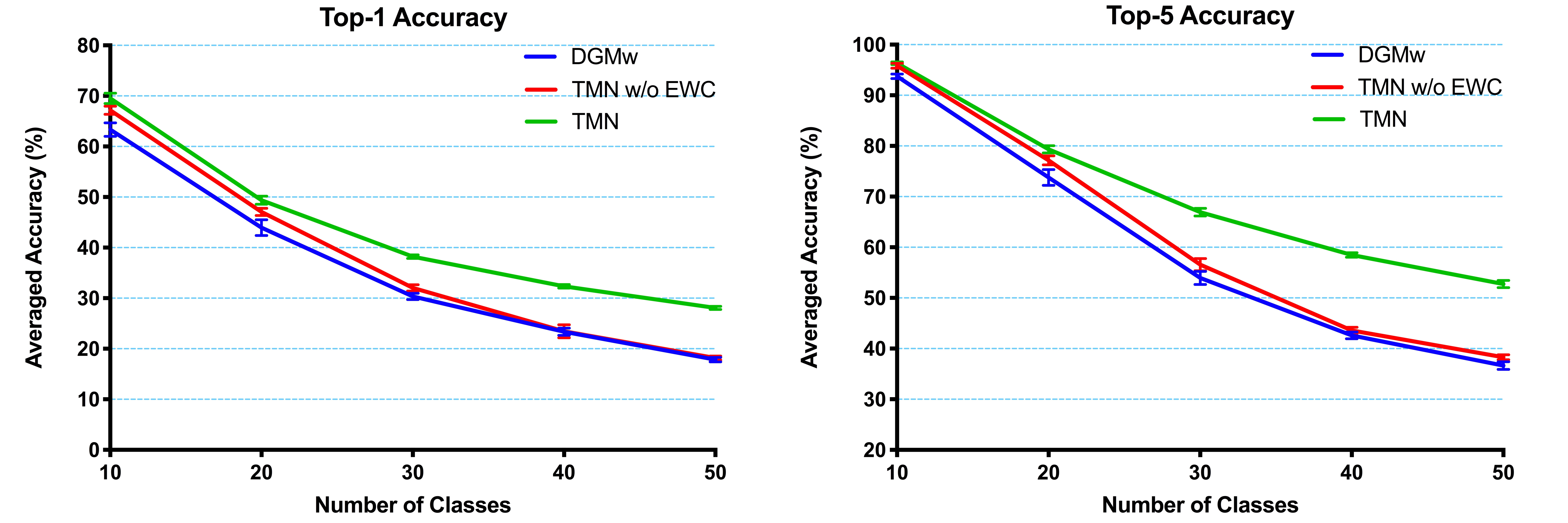} 
     \vspace{-.1cm}
	\caption{Averaged top-1 and top-5 accuracy of class-incremental learning on ImageNet.}
	\label{fig3}
    \vspace{-.2cm}
\end{figure}

\begin{table}[ht]
    \vspace{-.2cm}
	\centering
	\caption{Averaged cosine similarity of empirical FIM of ResNet18 last two blocks in the 10-class ImageNet task. We train the classifier with training data first and then generated data of the same task under different strength of weight consolidation (\(\lambda_{C}\)).}\smallskip
     \vspace{-.2cm}
	\resizebox{0.9\columnwidth}{!}{
	\begin{tabular}{ccccccccc}
		\hline
       \multicolumn{1}{c}{}& \multicolumn{4}{c}{block 3} & \multicolumn{4}{c}{block 4} \\
		Parameter & \(\lambda_{C}=0\) & \(\lambda_{C}=10\)&\(\lambda_{C}=100\)&  \(\lambda_{C}=1000\) & \(\lambda_{C}=0\) & \(\lambda_{C}=10\)&\(\lambda_{C}=100\)&  \(\lambda_{C}=1000\)\\
		\hline
		0.layers.0.weight&0.8294&0.8089&0.9214&0.9495&0.4571&0.8673&0.9209&0.8704\\
		0.layers.1.weight&0.8350&0.9371&0.9653&0.9943&0.6411&0.8460&0.9120&0.9899\\
		0.layers.3.weight&0.6767&0.8110&0.9184&0.9519&0.6364&0.7749&0.8928&0.8309\\
		0.layers.4.weight&0.6039&0.8101&0.9541&0.9501&0.6324&0.8549&0.8909&0.9059\\
		0.shortcut.0.weight&0.7766&0.8127&0.9199&0.8144&0.6451&0.7602&0.6837&0.8510\\
		0.shortcut.1.weight&0.8230&0.8953&0.9110&0.9626&0.5553&0.8444&0.8581&0.9302\\
		1.layers.0.weight&0.7934&0.8500&0.9478&0.9334&0.4826&0.8008&0.8754&0.9241\\
		1.layers.1.weight&0.8150&0.9840&0.8992&0.9988&0.5160&0.9152&0.9172&0.9693\\
		1.layers.3.weight&0.7361&0.9333&0.9083&0.9608&0.2159&0.8247&0.8331&0.8629\\
		1.layers.4.weight&0.6755&0.8904&0.9270&0.9828&0.2774&0.8747&0.9408&0.9776\\
		\hline
	\end{tabular}
	}
	\label{table3}
	\vspace{-.2cm}
\end{table}

Thirdly, we measure empirical FIM, i.e. squared gradient of the parameters, of a classifier trained on training data and then trained on generated data of the same task. Higher strength (larger \(\lambda_{C}\)) of weight consolidation increases similarity of FIM of parameters on training data and generated data during generative replay (Table 3). The directions of FIM on the generated data and the training data are much closer under higher strength of weight consolidation, which regularize optimization of the parameters on generated data to a closer direction as the training data.

\begin{table}[ht]
	\centering
	\caption{Averaged similarity of the empirical FIM in AC-GAN. We calculate the cosine similarity (Cosine) and correlation coefficient (Correlation) after 10-class incremental learning on MNIST or SVHN dataset.}\smallskip
    \vspace{-.2cm}
	\resizebox{0.8\columnwidth}{!}{
	\begin{tabular}{ccccc}
		\hline
		\multicolumn{1}{c}{}& \multicolumn{2}{c}{MNIST} & \multicolumn{2}{c}{SVHN} \\
		Parameter & Cosine & Correlation & Cosine & Correlation\\
		\hline
		conv1.weight&0.6586&0.2285&0.7000&-0.0196\\
		conv2.weight&0.3996&0.3566&0.2031&0.0806\\
		conv3.weight&0.1625&0.1486&0.0783&-0.0061\\
		BatchNorm2.weight&0.5078&0.0250&0.4378&-0.0869\\
		%BatchNorm2.bias&0.4722&0.0905&0.5232&-0.0598\\
		BatchNorm3.weight&0.5576&-0.0251&0.4788&-0.0747\\
		%BatchNorm3.bias&0.5787&-0.0300&0.5555&-0.0699\\
		\hline
	\end{tabular}
	}
	\label{table4}
	\vspace{-.1cm}
\end{table}

\subsection{Effectiveness of Triple-Network Architecture}
A key difference of TMNs from many single-head generative replay methods is the relatively independent classifier. Since \(D'\) can also make the prediction, we use a simple decision-making equation for the final prediction. Now we examine the necessity of the additional classifier and the decision-making process. One evidence has been mentioned that our preliminary experiment (Table 4) shows divergent directions of FIMs of the discriminator and the auxiliary classifier in the shared network on AC-GAN architecture, which interfere with each other.

We also quantify the classification results of individual classifiers and the final prediction after decision-making in Table 5. In all the experiments above, the averaged accuracy of the final prediction is always higher than individual D' and C. Moreover, TMNs w/o EWC in Table 2 and Fig. 3 outperforms DGMw, which uses the same form of conditional generator but on an AC-GAN architecture. Notably, the first data point \(A_{10}\) in Fig. 3 is the averaged accuracy of the first incremental batch in ImageNet-50 experiment. The first incremental batch only uses training data rather than generated data and there is no weight regularization. Both top-1 and top-5 \(A_{10}\) of TMNs (69.53, 96.32) significantly outperform DGMw (63.34, 93.75).  Thus, the triple-network architecture further alleviates catastrophc forgetting.

\begin{table}[h]
    \vspace{-.2cm}
	\caption{Averaged accuracy (\%) ($ \pm $SEM) of individual network and final output of TMNs. }\smallskip
    \vspace{-.2cm}
	\centering
	\resizebox{0.7\columnwidth}{!}{
		\smallskip\begin{tabular}{cccc}
		\hline
		 & D' & C & Output\\
		\hline
        SVHN $ A_{5} $& 86.34\,($ \pm $0.19)\,&86.35\,($ \pm $0.29)\,&\textbf{87.12\,($ \pm $0.22)}\\
        SVHN $ A_{10} $&74.81($ \pm $0.32)&75.12($ \pm $0.17)&\textbf{77.08($ \pm $0.26)}\\
        CIFAR-10 $ A_{5} $&70.40($ \pm $0.29)&69.36($ \pm $0.45)&\textbf{72.72($ \pm $0.36)}\\
        CIFAR-10 $ A_{10} $&59.96($ \pm $0.23)&53.11($ \pm $0.47)&\textbf{61.24($ \pm $0.14)}\\
        ImageNet-50 $ A_{30} $&36.46($ \pm $0.84)&36.65($ \pm $0.81)&\textbf{38.23($ \pm $0.75)}\\
        ImageNet-50 $ A_{50}$&26.90($ \pm $0.50)&24.33($ \pm $0.48)&\textbf{28.08($ \pm $0.33)}\\
		\hline
	\end{tabular}
}
	\label{table5}
	\vspace{-.2cm}
\end{table}

\section{Conclusions}
In this work, we analyze how brain memory system encodes, consolidates and complements specific and generalized information to successfully overcome catastrophic forgetting. Inspired by the organization principles of brain memory system, we apply a triple network architecture of GAN to model the interplay of hippocampus, prefrontal cortex and sensory cortex. Inspired by the neural mechanisms to consolidate specific or generalized information, we implement the three modules with appropriate brain-inspired algorithms to develop the ``artificial memory networks''. The triple-network architecture consists of a classifier network to show its advantages in classification, which can be replaced by a corresponding task-solver network and weight regularization methods to cope with other tasks. Further work will focus on more accurately modeling synaptic plasticity and extending the framework to other continual learning scenarios. 

\clearpage
\bibliographystyle{plainnat}
\bibliography{main}
\end{document}